\documentclass[lettersize,journal]{IEEEtran}

\usepackage{amsmath,amsfonts}

\usepackage{graphicx}

\usepackage{booktabs}
\usepackage{multirow}
\usepackage{array}

\usepackage[table]{xcolor}

\usepackage[caption=false,font=normalsize,labelfont=sf,textfont=sf]{subfig}

\usepackage{algorithmic}
\usepackage{algorithm}

\usepackage{textcomp}
\usepackage{stfloats}
\usepackage{url}
\usepackage{verbatim}

\usepackage{cite}

\usepackage{microtype}

\usepackage[colorlinks,allcolors=blue]{hyperref}

\begin{document}

\title{DDA-Thinker: Decoupled Dual-Atomic Reinforcement Learning for Reasoning-Driven Image Editing}

\author{
    Hanqing Yang,
    Qiang Zhou,
    Yongchao Du,
    Sashuai Zhou,
    Zhibin Wang,
    \\
    Jun Song,
    Tiezheng Ge,
    Cheng Yu,
    Bo Zheng
    \thanks{Hanqing Yang, Qiang Zhou, Yongchao Du, Sashuai Zhou, Zhibin Wang, Jun Song, Tiezheng Ge, Cheng Yu, and Bo Zheng are with Alibaba Group. Sashuai Zhou is also with Zhejiang University.}
    \thanks{Hanqing Yang, Qiang Zhou, Yongchao Du, and Sashuai Zhou contributed equally to this work.}
    \thanks{Jun Song is the corresponding author.}
}

\maketitle

\begin{abstract}
Recent image editing models have achieved strong visual fidelity but often struggle with tasks requiring complex reasoning. To investigate and enhance the reasoning-grounded planning for image editing, we propose \textbf{DDA-Thinker}, a Thinker-centric framework designed for the independent optimization of a planning module (Thinker) over a fixed generative model (Editor). This decoupled Thinker-centric paradigm facilitates a controlled analysis of the planning module and makes its contribution under a fixed Editor easier to assess. To effectively guide this Thinker, we introduce a dual-atomic reinforcement learning framework. This framework decomposes feedback into two distinct atomic rewards implemented through verifiable checklists: a cognitive-atomic reward to directly assess the quality of the Thinker’s executable plan, which serves as the actionable outcome of the Thinker’s reasoning, and a visual-atomic reward to assess the final image quality. To improve checklist quality, our checklist synthesis is grounded not only in the source image and user instruction but also in a rational reference description of the ideal post-edit scene. To support this training, we further develop a two-stage data curation pipeline that first synthesizes a diverse and reasoning-focused dataset, then applies difficulty-aware refinement to curate an effective training curriculum for reinforcement learning. Extensive experiments on reasoning-driven image editing benchmarks, including RISE-Bench and KRIS-Bench, demonstrate that our approach substantially improves overall performance. Our method enables a community model to achieve results competitive with strong proprietary models, highlighting the practical potential of Thinker-centric optimization under a fixed-editor setting.
\end{abstract}


\section{Introduction}

\begin{figure}[t]
\centering
\includegraphics[width=\linewidth]{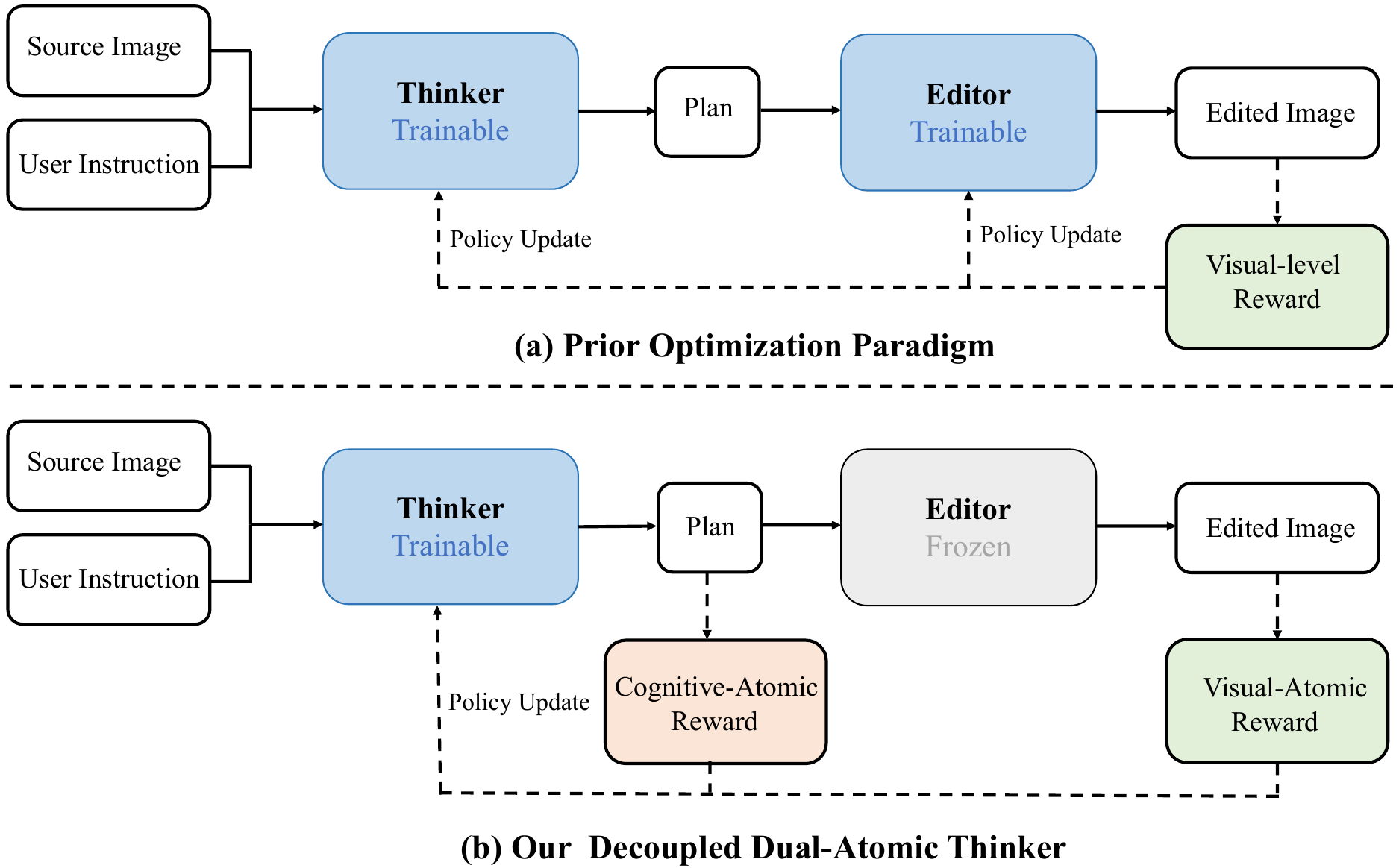}
\caption{\textbf{Comparison of Training Paradigms.} 
\textbf{(a) Prior paradigms} jointly or alternately train both Thinker and Editor. In this intertwined optimization, feedback from a single outcome is used to update both modules, making it harder to isolate whether errors arise from planning or execution. Additionally, visual-only rewards may provide limited guidance on the quality of the executable plan.
\textbf{(b) Our DDA-Thinker} decouples planning from generation by exclusively optimizing the Thinker while keeping the Editor frozen. 
This design decouples the Thinker's optimization, allowing its contribution to be more easily assessed and guided with fine-grained dual-atomic feedback on both the executable plan and the visual outcome.
}
\label{fig:paradigm_comparison} 
\end{figure}

Recent advances in image editing have been significantly propelled by large-scale vision-language models and diffusion models~\cite{qwenimage-edit, blackforest2024flux, longcat,dhariwal2021diffusion,wallace2024diffusiondpo,DBLP:journals/pami/HuangHLYLLXZCC25,DBLP:journals/pami/freeedit}. These models demonstrate strong capabilities in attribute-level modifications~\cite{brooks2023instructpix2pix,DBLP:journals/pami/WengGZWLS25,liu2025step1x-edit,liao2025imagegen-cot,DBLP:journals/pami/HuangDSXLL25} such as color adjustment, object insertion, and style transfer. However, despite their impressive visual fidelity, they often struggle with tasks requiring complex reasoning, such as maintaining physical plausibility, logical coherence, and factual correctness~\cite{lian2024llmgroundeddiffusionenhancingprompt,li2025editthinker,zhou2026spatialreward,wang2024t2i_compbench,risebench}. This limitation largely stems from the absence of an explicit planning stage: these systems directly translate user instructions into visual modifications without reasoning about the underlying physical, causal, or logical constraints that govern the desired transformation.

To bridge this gap, recent works have explored incorporating explicit planning into editing pipelines~\cite{unifiedthinker, thinkrl, thinkgen, li2025editthinker, yin2025reasonedit}. For example, Unified Thinker~\cite{unifiedthinker} and ThinkRL-Edit \cite{thinkrl} introduce planning components to bridge user instructions and image editing operations. While promising, these methods typically rely on optimization strategies where planning and generation components are jointly or alternately trained. As illustrated in Figure~\ref{fig:paradigm_comparison}(a), prior paradigms raise two practical issues for analyzing and improving the planning module. First, the intertwined optimization complicates error attribution, as a poor edit may arise either from a flawed plan or from the generator’s failure to execute an otherwise valid one. As a result, the contribution of the planning module becomes harder to assess cleanly, motivating our decoupled Thinker-centric setting for more controlled investigation. Second, many existing reinforcement learning schemes in this domain \cite{thinkrl,unifiedthinker,thinkgen} largely depend on visual-only reward signals. These signals are often unreliable due to coarse-grained assessment and offer only indirect evidence about the underlying executable plan. This creates a ``cognitive blind spot": a flawed executable plan can still be positively reinforced if the final image appears acceptable, while a sound plan paired with imperfect generation is penalized. As a result, the policy may over-rely on visually plausible outcomes even when the underlying executable plan is incomplete or imprecise.

In this work, we explore an alternative to the prevailing paradigms where both planning and generation modules are actively trained. We shift our focus toward a more controlled investigation, asking a different research question: \textit{\textbf{To what extent can performance in reasoning-intensive image editing be improved by exclusively optimizing the planning module?}} This question is motivated by our insight that for a fixed editor, the quality of the executable plan often represents a key immediate and tractable bottleneck. Here, planning refers to the form in which the Thinker turns its reasoning into instructions executable by the Editor. To answer this question, we propose DDA-Thinker, a framework built upon a Thinker-centric paradigm (Figure~\ref{fig:paradigm_comparison}(b)). A dedicated planning module (Thinker) is trained for a strictly frozen image editor. We clarify that our goal is not to establish the superiority of decoupled Thinker-centric optimization over joint or alternating training of the Thinker and Editor. Instead, we adopt a decoupled Thinker-centric setup as a controlled analytical setting to study the impact of improving the planning module under a fixed editor. Our approach therefore explores a Thinker-centric optimization setting: we focus on teaching the Thinker to produce plans that are logically sound and better aligned with the Editor’s execution capabilities.

However, such a decoupled Thinker-centric strategy introduces a critical challenge: how to provide effective and fine-grained feedback for the Thinker. Existing reward formulations ranging from scalar scores to visual-only checklists~\cite{thinkrl, unifiedthinker, flowgrpo, dancegrpo} are often inadequate, as they either fail to capture the quality of the executable plan or are susceptible to ambiguity and noise. To overcome these limitations, we propose a \textit{\textbf{dual-atomic reinforcement learning framework}}. This framework decomposes the reward into two distinct atomic rewards, each implemented through a verifiable checklist, for executable planning and visual outcomes: a cognitive-atomic reward assessing the executable planning output, which serves as the actionable outcome of the Thinker’s reasoning, in terms of intent preservation, logical soundness, and plan executability; and a complementary visual-atomic reward assessing the final image's instruction following, appearance consistency, and hallucination detection. We apply this cognitive-atomic reward at the plan level because the executable plan is the most direct and least ambiguous optimization target under a frozen Editor. To make these binary-verified checklists more precise and reliable, they are synthesized using not only the source image and user instruction but also a rational reference description, i.e., a textual depiction of the ideal post-edit scene. Grounding checklist generation in this textual reference helps reduce reward noise and ambiguity. By providing explicit guidance on both plan quality and visual outcomes, this dual-faceted feedback mechanism helps optimize the Thinker for more reliable execution.

Effective reinforcement learning under this framework requires a data corpus that is both diverse in reasoning scenarios and appropriately challenging for policy optimization. To this end, we develop a two-stage generative data pipeline. The first stage synthesizes a diverse reasoning-focused dataset for initial supervised training: guided by a reasoning-centric taxonomy, a powerful Large Language Model (LLM) acts as a scenario generator, creating data triplets consisting of source image description, user instruction, and rational reference description. Source images are then synthesized from their corresponding descriptions using a high-fidelity text-to-image (T2I) model. This generation-driven strategy offers greater control over scenario distribution and scenario coherence, helping mitigate the coverage limitations often found in web-crawled or repurposed datasets. Subsequently, the second stage applies difficulty-aware refinement, curating a training curriculum by filtering out samples that are either too easy or too hard. This process yields a dataset that provides a stronger and more targeted learning signal, thereby supporting more effective policy optimization.

Our contributions are threefold:
\begin{itemize}
    \item 
    We propose a decoupled Thinker-centric paradigm that independently optimizes the planning module, enabling a controlled study of how improving the Thinker affects the final editing outcomes under a fixed Editor.
    \item 
    To effectively guide this paradigm, we introduce a dual-atomic reinforcement learning framework. It provides fine-grained verifiable feedback on both the executable plan and the visual outcome through checklist-based rewards, with checklist generation additionally grounded in rational reference descriptions.
    \item  
     Through extensive experiments on reasoning-driven image editing benchmarks, we show that Thinker-centric optimization can substantially improve a strong fixed Editor. Under this setting, our method establishes new state-of-the-art results among community models and achieves performance competitive with leading proprietary models.
\end{itemize}

\section{Related Work}

\subsection{Reasoning-Driven Image Editing}
General image editing models \cite{qwenimage-edit,journals/pami/XuCTJ22,tip-consistent,tip-Instruction,tip-generation,blackforest2024flux,DBLP:journals/pami/Talk-to-Edit,DBLP:journals/pami/KongLWLRK25} have achieved remarkable proficiency in basic attribute-level modifications but often fail to handle tasks requiring sophisticated physical causality or temporal evolution. To bridge this gap, recent works \cite{unifiedthinker,thinkrl,li2025editthinker,deepgen,unit,thinkgen,endocot}  have integrated planning modules into the editing pipeline. However, these methods often rely on complex coupled optimization strategies and frequently suffer from a lack of fine-grained reinforcement feedback during training. Specifically, ThinkGen \cite{thinkgen} employs a training pipeline that requires an auxiliary instruction refinement module and per-scenario rule-based models. Unified Thinker \cite{unifiedthinker} relies on joint supervised fine-tuning of both planning and generative components, which complicates the optimization landscape and limits feedback granularity. ThinkRL-Edit \cite{thinkrl} employs a training paradigm that involves updating both the planning and generation modules. Its feedback mechanism focuses predominantly on image-level verification, largely neglecting the cognitive plan itself. While effective, these optimization strategies complicate the assessment of the planning module’s contribution to the final editing outcome. Furthermore, iterative architectures such as EditThinker \cite{li2025editthinker} and ReasonEdit \cite{yin2025reasonedit} utilize multi-round reflection cycles to improve accuracy, yet such designs incur substantial computational overhead and inference latency.

In contrast, we adopt a Thinker-centric strategy that provides a more controlled setting for studying the planning module. Specifically, we optimize the Thinker while keeping the generative model strictly frozen, thereby decoupling planning optimization from image generation. By introducing a dual-atomic reinforcement learning framework, our approach provides fine-grained reward signals to better align planning quality with visual execution.

\subsection{Reinforcement Learning for Image Editing}
Recent research \cite{flowgrpo,promptrl,journals/tip/LinLLLFH24,replan, unireason,think_then_generate,edit-r1} has increasingly employed reinforcement learning to improve the capabilities of Vision-Language Models (VLMs) and diffusion-based models. A core challenge in this domain is translating holistic evaluation into actionable training signals for image editing, where success depends on both accurate intent comprehension and high-precision pixel-level execution~\cite{wisebench,clipreason,onereward,DBLP:journals/tip/LinW23,risebench}. Several existing approaches \cite{editscore,editreward} prioritize the development of domain-specialized reward models to improve the quality assessment of edited images. For example, EditScore \cite{editscore} trains specialized VLM-based judges to provide more reliable feedback than general-purpose reward models. Other methods \cite{unifiedthinker, thinkrl, thinkgen} leverage pre-trained foundation models as reward models to generate training signals for optimizing both the reasoning VLM and the diffusion-based generation model. Some of these methods have introduced checklist-style rewards to decompose complex visual editing tasks into measurable units. However, they remain limited to visual-level signals and do not provide direct feedback on the executable plan. Furthermore, the checklist generation process in such methods often lacks grounding in the intended editing result, which can lead to noisy rewards.

In this work, we propose a dual-atomic reinforcement learning framework to address these limitations. Rather than relying on scalar-based or purely visual feedback, our approach decomposes the reward process into binary-verified checklists across both visual and cognitive domains. To ensure the quality and reliability of these checklists, their generation is guided not only by the source image and user instruction, but also by a rational reference description of the ideal post-edit scene. This description serves as a semantic reference, helping align checklist generation with the intended editing target and reduce potential noise and ambiguity.

\section{Methodology}
\label{sec:method}
\begin{figure*}[t]
\centering
\includegraphics[width=\textwidth]{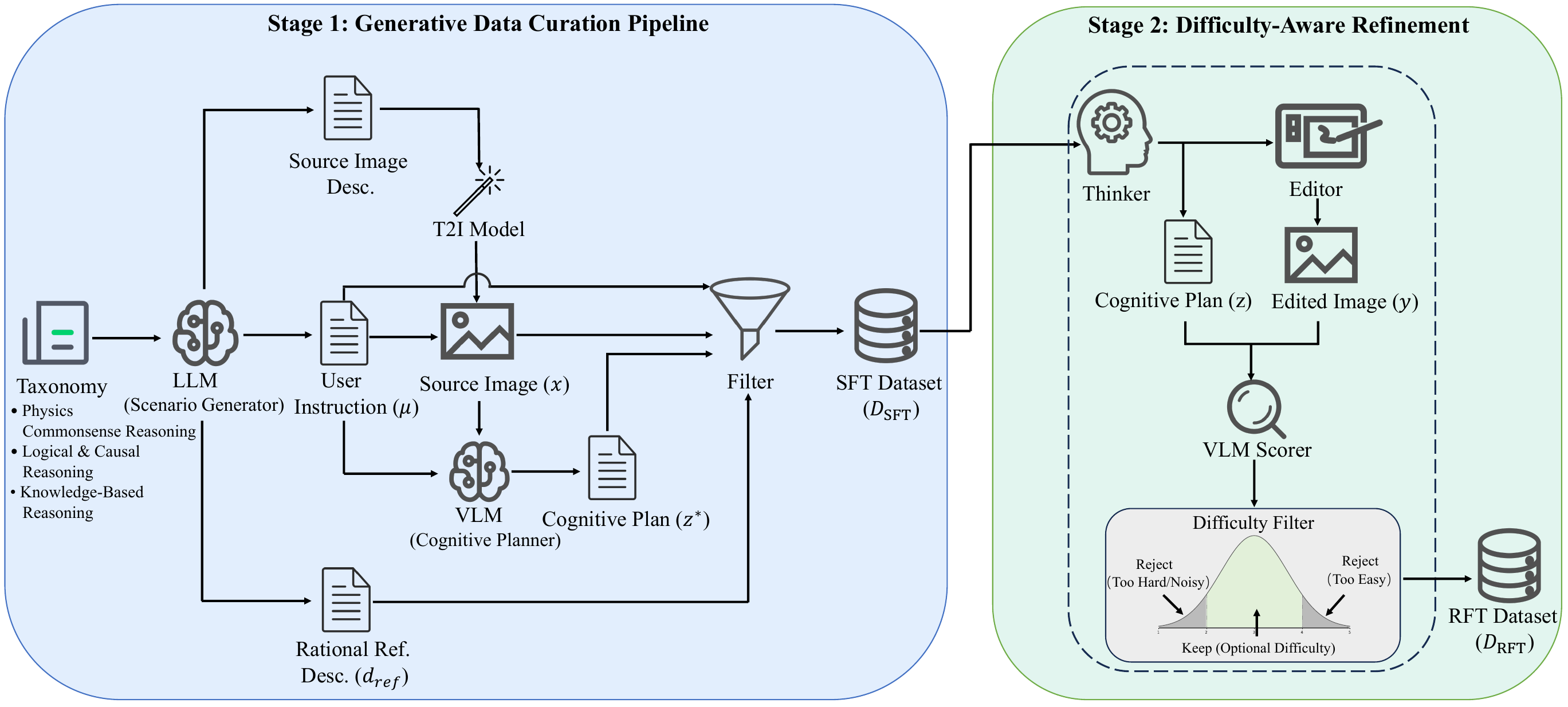}
\caption{\textbf{Overview of the Reasoning-Enhanced Data Curation Process.} This process constructs two datasets: an initial set for supervised fine-tuning ($\mathcal{D}_\text{SFT}$) and a refined set for reinforcement learning ($\mathcal{D}_\text{RFT}$). \textbf{Stage 1 (Generative Data Curation Pipeline):} Guided by a reasoning taxonomy, an LLM serves as the scenario generator to create diverse data triplets. A T2I model then synthesizes the source images, and a VLM-based planner produces the corresponding cognitive plans. This stage is designed to achieve broad scenario coverage and yields the initial dataset $\mathcal{D}_\text{SFT}$.
\textbf{Stage 2 (Difficulty-Aware Refinement):} Samples from $\mathcal{D}_\text{SFT}$ are processed by our Thinker and Editor. A VLM-based evaluator then scores the quality of the overall outcome. Based on this score, a difficulty filter selects samples within a moderate difficulty range, creating the final dataset $\mathcal{D}_\text{RFT}$ by rejecting those that are too easy or too hard/noisy.
}
\label{fig:data_curation}
\end{figure*}

In this work, we propose \textbf{DDA-Thinker}, a framework for reasoning-driven image editing. Our approach adopts a \textbf{Thinker-centric strategy}, where we exclusively optimize a Thinker module to act as an intelligent planner for a frozen Editor. This strategy is realized through a dual-atomic reinforcement learning framework, which refines the Thinker's planning abilities by providing fine-grained feedback on both the executable plan and visual outcomes. The entire training process is supported by a two-stage generative data pipeline designed to provide a diverse and effective training curriculum. This section first outlines our conceptual framework, then details the data synthesis and training processes that enable our Thinker-centric approach.

\subsection{Conceptual Framework and Innovations}
\label{ssec:innovations}
While prior works often employ schemes where both planning and generation modules are actively trained \cite{unifiedthinker, thinkrl, thinkgen}, our approach is intentionally designed to optimize the Thinker module in a more controlled setting. We clarify that our goal is not to establish the superiority of this Thinker-centric strategy over joint or alternating training of the Thinker and Editor. This methodology is motivated by the observation that a sufficiently capable Thinker can learn to better leverage the capabilities of a fixed generative model. To this end, we adopt a Thinker-centric strategy, concentrating all optimization efforts on the Thinker while treating the Editor as a frozen execution engine throughout training.

However, this Thinker-centric strategy introduces a critical challenge: how to provide a reward signal that is both fine-grained and reliable enough to guide the Thinker effectively. Conventional reward signals, such as interval-based scalar scores or visual-only feedback, are often insufficient for this purpose. To address this, we build upon the principle of atomic rewards, which decompose a single ambiguous score~\cite{unifiedthinker,flowgrpo,dancegrpo,decomposeAC} into a set of verifiable questions, each yielding a binary outcome. Building on this principle, our dual-atomic reinforcement learning framework advances beyond the visual-only feedback schemes by addressing two key limitations. First, to improve checklist quality and reduce noise, our checklist generation is grounded not only in the source image and user instruction but also in a rational reference description of the ideal post-edit scene. Second, we introduce a cognitive-atomic reward that explicitly evaluates the quality of the Thinker’s executable plan. This dual-faceted feedback, supported by a purpose-built data corpus, effectively helps optimize the Thinker for stronger planning and more reliable visual outcomes.

\subsection{Reasoning-Enhanced Data Corpus}
\label{ssec:data_synthesis}
Effective training of the dedicated Thinker for reasoning-driven image editing requires a data corpus with diverse reasoning-driven scenarios and an appropriate level of task difficulty. To this end, we develop a two-stage data curation pipeline, as illustrated in Figure~\ref{fig:data_curation}.

\subsubsection{Generative Data Curation Pipeline}
Our generative data curation pipeline, depicted in Stage 1 of Figure~\ref{fig:data_curation}, allows us to better control scenario content during data construction. To build a training data corpus that covers diverse reasoning demands in image editing, we guide generation with a taxonomy spanning three broad domains. The first, Physical Commonsense Reasoning, includes scenarios requiring reasoning about material properties, object interactions, temporal evolution, and spatial relationships. The second, Logical \& Causal Reasoning, focuses on the ability to deduce and infer abstract relationships, eliciting tasks involving multi-step problem solving, causal inference, and adherence to logical constraints. The third, Knowledge-Based Reasoning, includes edits that depend on domain-specific knowledge, such as historical facts, cultural context, and biological processes. Guided by this taxonomy, a powerful off-the-shelf LLM serves as the scenario generator to create data triplets consisting of source image description, user instruction, rational reference description. The rational reference description, a textual depiction of the ideal post-edit scene, is specifically introduced to support the later generation of more reliable checklists. Source images are then synthesized from their corresponding descriptions using a high-fidelity T2I model. This generative strategy helps mitigate the coverage limitations often found in web-crawled or repurposed datasets, where reasoning-intensive editing scenarios are often scarce \cite{unifiedthinker, thinkgen, editreward}. Subsequently, for each sample, a powerful VLM-based planner is employed to synthesize the cognitive plan in a structured format:
$z = \text{\texttt{<think>} } c \text{ \texttt{</think> <answer>} } a \text{ \texttt{</answer>}}$
where $c$ denotes the intermediate chain of thought and $a$ is the final executable plan. Finally, after automated checks and manual inspection, the resulting data forms our initial supervised training set $\mathcal{D}_{\text{SFT}}$.

\begin{figure*}[t]
\centering
\includegraphics[width=\textwidth]{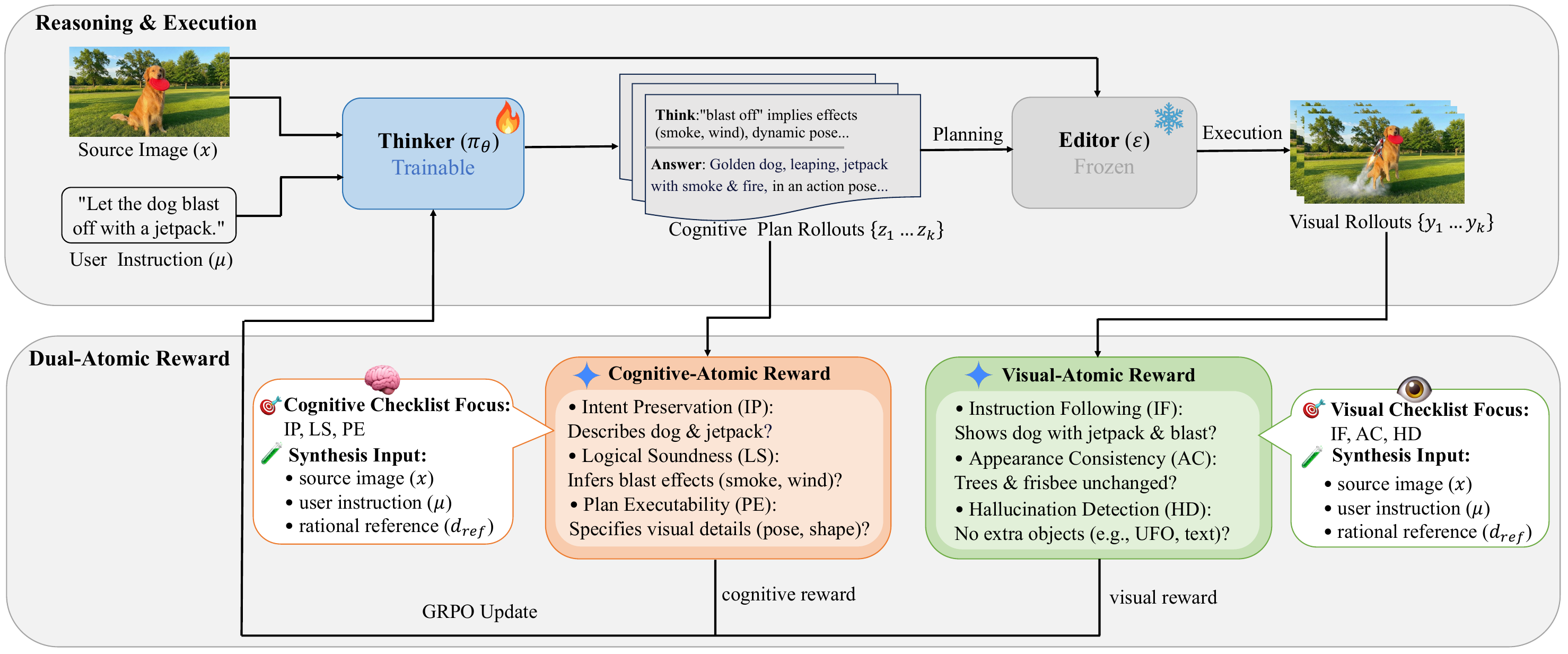}
\caption{\textbf{The DDA-Thinker Training Framework.} Our framework trains a Thinker ($\pi_{\theta}$) over a frozen Editor ($\mathcal{E}$). The core of our method is the dual-atomic reward mechanism, which provides fine-grained feedback on both the executable plan and the final image.
The \textbf{cognitive-atomic reward} assesses the plan's quality (e.g., IP, LS, PE), while the \textbf{visual-atomic reward} assesses the final image's quality (e.g., IF, AC, HD). The binary checklists for both rewards are synthesized using the source image ($x$), user instruction ($u$), and the rational reference description ($d_{\text{ref}}$). The combined rewards then update only the Thinker's policy via GRPO.}
\label{fig:framework}
\end{figure*}

\subsubsection{Difficulty-Aware Refinement}
To construct a more effective training curriculum for the reinforcement learning stage, we perform difficulty-aware rejection sampling, as depicted in Stage 2 of Figure~\ref{fig:data_curation}. This strategy is motivated by the observation that not all samples provide equally useful learning signals \cite{minimalist}. Trivial examples offer diminishing returns, while excessively difficult or noisy ones, where the model consistently fails, can introduce misleading gradients and hinder policy updates \cite{minimalist}. Our goal is therefore to curate a dataset of moderately difficult samples, which represent an effective ``sweet spot" for learning \cite{minimalist,Star-DS}. Specifically, we employ a powerful VLM-based evaluator to provide a holistic quality score for each attempted edit. This score serves as a heuristic proxy for overall sample quality, informed by two key aspects: the quality of the generated plan (e.g., clarity, correctness, executability) and the visual quality of the final image (e.g., instruction following, consistency, plausibility). We then filter out samples that are either too easy (high score) or too hard/noisy (low score), retaining only those within a moderate difficulty range. This curated set, denoted as $\mathcal{D}_{\text{RFT}}$, provides a more robust and sample-efficient curriculum for subsequent policy optimization.

\subsection{Thinker-Centric Dual-Atomic Optimization}
\label{ssec:training}
Our training procedure is designed to exclusively optimize the Thinker's policy $\pi_\theta(z \mid x, u)$, parameterized by $\theta$, which generates a structured cognitive plan $z$ based on the source image $x$ and user instruction $u$. Throughout this process, the Editor $\mathcal{E}(x, a)$ remains frozen. The training unfolds in two main phases.
  
\subsubsection{Supervised Fine-Tuning Initialization}
We first initialize the Thinker's policy $\pi_\theta$ via Supervised Fine-Tuning (SFT). Given the training dataset $\mathcal{D}_{\text{SFT}}$ of ground-truth triples $(x, u, z^\ast)$, the objective is to minimize the negative log-likelihood of the expert-written target $z^\ast$:
\begin{equation}
\mathcal{L}_{\text{SFT}} = - \mathbb{E}_{(x,u,z^\ast)\sim \mathcal{D}_{\text{SFT}}} \left[ \log \pi_\theta(z^\ast \mid x,u) \right].
\end{equation}
This stage aligns the Thinker with the desired output format and teaches it to transform the user instruction into a final executable plan, accompanied by intermediate reasoning chain in the structured response.

\subsubsection{Reinforcement Fine-Tuning with Dual-Atomic Rewards}
After supervised initialization, we further optimize the Thinker using reinforcement learning. As illustrated in Figure~\ref{fig:framework}, the core of this stage is to move beyond a single ambiguous score by decomposing feedback into a set of atomic and verifiable questions. Our dual-atomic reward mechanism extends this principle by addressing two limitations in prior checklist-based methods. First, it improves the quality and reliability of the checklist questions. By grounding offline checklist generation not only in the source image and user instruction but also on a rational reference description of the ideal post-edit scene, our approach reduces noise in the generated checklist. Second, it expands the scope of feedback beyond visual reward by introducing a cognitive-atomic reward to directly guide the Thinker's executable plan.

We formalize the offline preparation of each checklist $Q$ (visual or cognitive) as sampling an optimal set of questions from a conditional probability distribution:
\begin{equation}
\label{eq:checklist_gen}
Q^* = \arg\max_{Q} P(Q \mid x, u, d_{\text{ref}}),
\end{equation}
where $x$ is the source image, $u$ is the user instruction, $d_{\text{ref}}$ is the rational reference description, and $P(Q \mid \cdot)$ represents the likelihood of a question set being effective and comprehensive. In practice, we approximate this ideal by prompting a powerful VLM to generate a high-quality checklist $Q$.

\vspace{1ex}
\noindent\textbf{Visual-Atomic Reward.} 
This reward assesses the overall quality and correctness of the edited output image $y$, providing a visual feedback signal that encourages the Thinker to generate plans leading to better visual outcomes. To generate a high-quality checklist for this assessment, we leverage the rational reference description, i.e., a textual depiction of the expected post-edit scene. Rather than relying on an explicit post-edit reference image, which is often difficult to construct reliably, this approach uses the description as a semantic reference. Grounded in this reference, together with the source image and user instruction, we generate a set of $M$ visual-related binary questions $Q^{\text{visual}}$. The visual checklist is designed to primarily probe three core attributes: (1) Instruction Following (IF), which assesses whether the primary edits are correctly applied; (2) Appearance Consistency (AC), which verifies that unedited regions remain unchanged; (3) Hallucination Detection (HD), which ensures that no extraneous objects or artifacts are introduced. A VLM-based reward model, denoted as $\mathcal{R}(\cdot)$, conceptually provides a binary score for each question. In practice, for efficiency, all $M$ questions for a given sample are evaluated in a single forward pass. The resulting visual reward is the average success rate, formally defined as:
\begin{equation}
r_{\text{visual}} = \frac{1}{M}\sum_{i=1}^{M} \mathcal{R}(x, u, y, q_i^{\text{visual}}),
\end{equation}
where $x$ is the source image, $u$ is the user instruction, $y$ is the edited output image being evaluated, $q_i^{\text{visual}} \in Q^{\text{visual}}$ is the specific question, and the output of $\mathcal{R}(\cdot)$ lies in $\{0,1\}$.

\vspace{1ex}
\noindent\textbf{Cognitive-Atomic Reward.} 
A visual-level reward is essential, but it provides only indirect feedback for the Thinker’s planning process. A good image may be generated from an imprecise plan, offering a noisy learning signal for the Thinker's planning process. To address this, our cognitive-atomic reward directly assesses the quality of the Thinker's executable plan, independent of the final visual outcome. The cognitive-atomic reward is calculated based exclusively on the final executable plan $a$ within the \texttt{<answer>} field. This design choice is directly aligned with our core objective: to optimize the quality of the actionable instruction sent to the Editor. The final plan $a$ serves as the most direct and least ambiguous proxy for this objective. While the \texttt{<think>} field contains the detailed reasoning process, our goal is to reward the successful outcome of that process, which is a correct and executable plan, since this is what ultimately determines the quality of the final image edit. Similar to its visual counterpart, the cognitive checklist $Q^{\text{cognitive}}$ is generated based on the source image, user instruction, and rational reference description. However, its purpose is fundamentally different: to probe the logical and semantic integrity of the executable plan. The cognitive checklist is designed to assess three key aspects of the cognitive plan:
(1) Intent Preservation (IP), which ensures that the plan accurately reflects the user's core request without omission or misinterpretation;
(2) Logical Soundness (LS), which verifies that the plan is factually correct and logically coherent, without violating commonsense;
(3) Plan Executability (PE), which confirms that the plan translates abstract concepts into actionable, concrete, and unambiguous instructions that the Editor can directly process. We employ the same VLM-based reward model $\mathcal{R}(\cdot)$ to evaluate the executable plan $a$, again processing all $N$ cognitive questions in a single forward pass for computational efficiency. The resulting cognitive reward is computed as:
\begin{equation}
r_{\text{cognitive}} = \frac{1}{N}\sum_{j=1}^{N} \mathcal{R}(x, u, a, q_j^{\text{cognitive}}),
\end{equation}
where $x$ is the source image, $u$ is the user instruction, $a$ is the executable plan being evaluated, $q_j^{\text{cognitive}} \in Q^{\text{cognitive}}$ is the specific question, and the output of $\mathcal{R}(\cdot)$  lies in $\{0,1\}$.

\vspace{1ex}
\noindent\textbf{Policy Optimization.}
Our dual-atomic framework provides two distinct reward signals, $r_{\text{cognitive}}$ and $r_{\text{visual}}$, which guide the Thinker towards plan quality and visual effectiveness, respectively. A common approach is to combine them into a single weighted sum and apply an optimization algorithm such as Group Relative Policy Optimization (GRPO)~\cite{shao2024deepseekmath} to the merged reward. However, this method entangles the two objectives through a shared advantage calculation and requires hand-tuning the weighting coefficient. We instead keep the two reward signals separate throughout optimization. During Reinforcement Fine-Tuning (RFT), every training example is used to form two parallel mini-batch streams, $\mathcal{D}_{\text{cog}}$ and $\mathcal{D}_{\text{vis}}$. This allows the same input to contribute to both objectives through independent rollout groups, rather than through a merged scalar reward. Both streams therefore cover the full training pool, meaning the two objectives are separated through distinct reward signals rather than by a disjoint data partition. For a mini-batch drawn from a specific stream $\mathcal{D}^* \in \{\mathcal{D}_{\text{cog}}, \mathcal{D}_{\text{vis}}\}$, the Thinker's policy $\pi_\theta$ is optimized using only the corresponding reward signal, so that it receives clear and focused feedback on each objective. The policy is updated using the GRPO objective:
\begin{equation}
\mathcal{L}_{\text{GRPO}} = - \mathbb{E}_{(x,u) \sim \mathcal{D}^*} \left[ \frac{1}{K} \sum_{k=1}^{K} A_k \log \pi_\theta(z_k\mid x,u) \right],
\end{equation}
where for each input $(x,u)$ we sample $K$ responses $\{z_1, \dots, z_K\}$ and score each response $z_k$ with the reward signal associated with $\mathcal{D}^*$. The advantage $A_k$ is the within-group standardized reward of $z_k$ computed over all $K$ responses. Since this standardization is performed independently for each reward channel, the resulting advantages are comparably scaled, thus requiring no hand-tuned weighting between them. By optimizing the Thinker with this dual-faceted and group-relative feedback, DDA-Thinker learns to generate executable plans that are not only logically sound but also highly executable, leading to high-quality visual edits.

\begin{table*}[t]
\centering
\caption{
Quantitative comparison on RISE-Bench across diverse reasoning dimensions. Performance is evaluated based on Instruction Reasoning (Reason.), Appearance Consistency (Consist.), and Visual Plausibility (Visual.). Accuracy (\%) is broken down into four core reasoning categories, and Overall Acc. is calculated as the mean of these four categories.
}
\label{tab:main results risebench}
\setlength{\tabcolsep}{8pt}
\fontsize{8pt}{12pt}\selectfont
\begin{tabular}{l|ccc|ccccc}
\toprule
\textbf{Model} & \textbf{Reason.} & \textbf{Consist.} & \textbf{Visual.} & \textbf{Temporal} & \textbf{Causal} & \textbf{Spatial} & \textbf{Logical} & \textbf{Overall Acc.} \\
\midrule

\rowcolor[HTML]{E8EAF6}
\multicolumn{9}{c}{\textit{Proprietary Models}} \\ 
GPT-Image-1-mini           & 54.1 & 71.5 & 93.7 & 24.7 & 28.9 & 33.0 &  9.4 & 24.4 \\
GPT-Image-1                & 62.8 & 80.2 & 94.9 & 34.1 & 32.2 & 37.0 & 10.6 & 28.9 \\
Gemini-2.5-Flash-Image (NanoBanana)     & 61.2 & 86.0 & 91.3 & 25.9 & 47.8 & 37.0 & 18.8 & 32.8 \\
Gemini-3-Pro-Image (NanoBanana-Pro) & 77.0 & 85.5 & 94.4 & 41.2 & 61.1 & 48.0 & 37.6 & 47.2 \\
\midrule

\rowcolor[HTML]{E0F2F1}
\multicolumn{9}{c}{\textit{Community Models}} \\
EMU2 \cite{emu2}                   & 22.6 & 38.2 & 78.3 &  1.2 &  1.1 &  0.0 &  0.0 &  0.5 \\
OmniGen \cite{xiao2024omnigen}                & 25.1 & 41.5 & 73.5 &  1.2 &  1.0 &  0.0 &  1.2 &  0.8 \\
Step1X-Edit  \cite{liu2025step1x-edit}         & 30.3 & 12.6 & 74.9 &  0.0 &  2.2 &  2.0 &  3.5 &  1.9 \\
Ovis-U1 \cite{ovis}               & 33.9 & 52.7 & 72.9 &  1.2 &  3.3 &  4.0 &  2.4 &  2.8 \\
FLUX.1-Kontext-Dev \cite{blackforest2024flux} & 26.0 & 71.6 & 85.2 &  2.3 &  5.5 & 13.0 &  1.2 &  5.8 \\
BAGEL (w/ CoT) \cite{bagel2025}        & 45.9 & 73.8 & 80.1 &  5.9 & 17.8 & 21.0 &  1.2 & 11.9 \\
Uni-CoT  \cite{unicot}       &  - & -  &  - & 8.2  & 18.9 &  20.0 & 1.2 & 12.5 \\
ThinkGen  \cite{thinkgen}       &  - & -  &  - & 16.4  & 17.7 &  16.0 & 1.1 & 13.0 \\
DeepGen  \cite{deepgen}       &  - & -  &  - & 15.3  & 18.9 &  14.0 & 4.7 & 13.3 \\
EditThinker \cite{li2025editthinker}  & - & - & -  & 10.8 & 23.3 &  27.0 & 8.2 & 17.8 \\
Unified Thinker \cite{unifiedthinker} & 61.9 & 76.2 & 90.5 &  32.9 & 30.0 & 41.0 &  9.4 & 28.9 \\
ThinkRL-Edit \cite{thinkrl} & 61.7 & 81.6 & 62.5 &  18.8 & 37.5 & 25.0 &  37.5 & 29.7 \\
\midrule

\rowcolor{gray!15}
Qwen-Image-Edit-2509 & 37.2 & 66.4 & 86.9  & 4.7 & 10.0 &  17.0 & 2.4 & 8.9 \\

\quad + DDA-Thinker-8B & 60.8 & 78.5 & 88.4 & 25.9 & 40.0 & 42.0 & 10.6& 30.3 \\

\quad + DDA-Thinker-32B & 67.2 & 79.2 & 89.5 & 34.1 & 36.7 & 47.0 & 15.1 & 33.9 \\

\rowcolor{gray!15}
Qwen-Image-Edit-2511 & 49.9 & 71.0 & 91.5  & 21.2 & 18.9 &  31.0 & 4.7 & 19.4 \\

\quad + DDA-Thinker-8B & 66.3 & 79.5 & 91.5 & 31.8 & 41.1 & 40.0 & 12.9 & 31.9 \\

\quad + DDA-Thinker-32B & \textbf{68.4} & \textbf{84.4} & \textbf{92.1} & \textbf{45.9} & \textbf{50.0} & \textbf{47.0} & \textbf{15.3} & \textbf{40.0} \\

\bottomrule
\end{tabular}
\end{table*}

\begin{table*}[t]
\centering
\caption{Performance evaluation on KRIS-Bench. Models are evaluated across three knowledge domains: Factual, Conceptual, and Procedural, further subdivided into seven dimensions. The Overall Score is computed as the average across these dimensions.}
\label{tab:main results krisbench}
\setlength{\tabcolsep}{3.5pt}
\fontsize{8.5pt}{10pt}\selectfont

\resizebox{\textwidth}{!}{
\begin{tabular}{l|cccc|ccc|ccc|c}
\toprule
\multirow{2}{*}{\textbf{Model}} & \multicolumn{4}{c|}{\textbf{Factual Knowledge}} & \multicolumn{3}{c|}{\textbf{Conceptual Knowledge}} & \multicolumn{3}{c|}{\textbf{Procedural Knowledge}} & \multirow{2}{*}{\textbf{Overall Average}} \\
\cmidrule(lr){2-5} \cmidrule(lr){6-8} \cmidrule(lr){9-11}
 & \textbf{Attribute} & \textbf{Spatial} & \textbf{Temporal} & \textbf{Average} & \textbf{Social Sci.} & \textbf{Natural Sci.} & \textbf{Average} & \textbf{Logical} & \textbf{Instruction} & \textbf{Average} & \\
\midrule

\rowcolor[HTML]{E8EAF6}
\multicolumn{12}{c}{\textit{Proprietary Models}} \\ 
Gemini 2.0       & 66.33 & 63.33 & 63.92 & 65.26 & 68.19 & 56.94 & 59.65 & 54.13 & 71.67 & 62.90 & 62.41 \\
Gemini-2.5-Flash-Image (NanoBanana)          & 78.97 & 80.17 &  86.49  & 80.48 &  83.7 & 76.34  &  78.11 & 63.96 & 90.44 & 75.31 & 78.17 \\
GPT-4o           & 83.17 & 79.08 & 68.25 & 79.80 & 85.50 & 80.06 & 81.37 & 71.56 & 85.08 & 78.32 & 80.09 \\
Gemini-3-Pro-Image  (NanoBanana-Pro)          & 85.61 & 87.17 &  88.06 &  86.36 & 85.25 & 83.64 & 84.03 & 81.42  &  93.22 & 86.48  & 85.31 \\
\midrule

\rowcolor[HTML]{E0F2F1}
\multicolumn{12}{c}{\textit{Community Models}} \\
InstructPix2Pix \cite{brooks2023instructpix2pix}  & 30.33 & 21.33 & - & 23.33 & 22.56 & 26.56 & 25.59 & 19.81 & 14.75 & 17.28 & 22.82 \\
OmniGen \cite{xiao2024omnigen}          & 37.92 & 28.25 & 21.83 & 33.11 & 30.63 & 27.19 & 28.02 & 11.94 & 35.83 & 23.89 & 28.85 \\
MagicBrush \cite{zhang2024magicbrush}       & 53.92 & 39.58 & - & 41.84 & 42.94 & 38.06 & 39.24 & 30.00* & 23.08 & 26.54 & 37.15 \\
AnyEdit \cite{yu2025anyedit}          & 47.67 & 45.17 & - & 39.26 & 38.56 & 42.94 & 41.88 & 36.56 & 26.92 & 31.74 & 38.55 \\
Emu2  \cite{emu2}            & 51.50 & 48.83 & 22.17 & 45.40 & 34.69 & 38.44 & 37.54 & 24.81 & 45.00 & 34.91 & 39.70 \\
Step1X-Edit \cite{liu2025step1x-edit}       & 55.50 & 51.75 & - & 45.52 & 44.69 & 49.06 & 48.01 & 40.88 & 22.75 & 31.82 & 43.29 \\
HiDream-E1 \cite{hidream}        & 52.75 & 49.42 & - & 43.31 & 52.56 & 49.25 & 50.05 & 45.19 & 30.08 & 37.64 & 44.72 \\
ByteMorph   \cite{bytemorph}      & 61.17 & 62.00 & - & 51.27 & 45.50 & 47.38 & 46.92 & 32.00 & 31.33 & 31.67 & 44.85 \\
FLUX.1-Kontext-Dev \cite{blackforest2024flux}   & 64.83 & 60.92 & - & 53.28 & 48.94 & 50.81 & 50.36 & 46.06 & 39.00 & 42.53 & 49.54 \\
OmniGen2  \cite{wu2025omnigen2}        & 59.92 & 52.25 & 54.75 & 57.36 & 47.56 & 43.12 & 44.20 & 32.50 & 63.08 & 47.79 & 49.71 \\
UniWorld-V1 \cite{lin2025uniworld}      & 58.17 & 54.50 & 63.00 & 47.71 & 47.50 & 43.94 & 44.80 & 42.00 & 53.83 & 47.92 & 50.27 \\
Step1X-Edit v1.1 \cite{liu2025step1x-edit}  & 64.17 & 61.75 & - & 53.05 & 52.06 & 55.06 & 54.34 & 52.56 & 36.75 & 44.66 & 51.59 \\
BAGEL  \cite{bagel2025}           & 64.27 & 62.42 & 42.45 & 60.26 & 55.40 & 56.01 & 55.86 & 52.54 & 50.56 & 51.69 & 56.21 \\
BAGEL-Think  \cite{bagel2025}     & 67.42 & 68.33 & 58.67 & 66.18 & 63.55 & 61.40 & 61.92 & 48.12 & 50.22 & 49.02 & 60.18 \\

Uni-CoT   \cite{unicot}        & 72.76 & 72.87 & 67.10 & 71.85 & 70.81 & 66.00 & 67.16 & 53.43 & 73.93 & 63.68 & 68.00 \\

ThinkRL-Edit  \cite{thinkrl}         & 81.02 & 81.45 & - & 81.13 & 75.67 & 71.25 & 72.31 & 49.07 & 79.71 & 57.44 & 71.65 \\
EditThinker   \cite{li2025editthinker}        & 78.48 & 73.83 & - & 77.24 & 76.20 & 70.69 & 72.02 & 65.23 & 66.89 & 65.94 & 71.91 \\

\midrule
\rowcolor{gray!15}
Qwen-Image-Edit-2509     & 72.44 & 79.10 & 55.07 & 71.05 & 66.14 & 57.75 & 59.78 & 49.83 & 77.61  & 61.94 & 63.71 \\

\quad + DDA-Thinker-8B   & 71.55 & 77.00 & 72.52 & 72.92 & 76.55 & 68.15 & 70.18 & 54.47 & 86.56 & 68.25 & 70.55 \\

\quad + DDA-Thinker-32B    & 72.79 & 77.92  &  73.09 & 73.98 & 76.90  & 70.60 & 72.13 & 55.43 & 89.39 & 70.07 & 72.21 \\

\rowcolor{gray!15}
Qwen-Image-Edit-2511      & 74.85 & 84.50    & 68.13 & 75.89 & 66.10 & 59.54 & 61.15 & 58.42 & 79.61 & 67.55 & 67.14 \\

\quad + DDA-Thinker-8B   & 79.14 & 79.33 & 74.43 & 78.41 & 79.86 & 76.93 & 77.64 & 68.97 & 80.11 & 73.83 & 76.99 \\

\quad + DDA-Thinker-32B          & \textbf{82.16} & \textbf{84.58} & \textbf{75.00} & \textbf{81.52} & \textbf{84.71} & \textbf{77.77} & \textbf{79.44} & \textbf{74.45} & \textbf{84.67} & \textbf{78.88} & \textbf{79.94} \\

\bottomrule
\end{tabular}
}
\end{table*}

\section{Experiments}

\subsection{Experimental Setup}

\vspace{1ex}
\noindent\textbf{Models and Training Setup.} 
We utilize Qwen3-VL-32B as the primary trained Thinker, with additional results reported for Qwen3-VL-8B \cite{bai2025qwen3vltechnicalreport}. Throughout training, the image generation Editor remains frozen. Unless otherwise specified, we use Qwen-Image-Edit-2511 \cite{qwenimage-edit} as the default Editor. The training pipeline follows a two-stage paradigm. First, the Thinker undergoes SFT on our 5k synthesized dataset, performed on 16 NVIDIA H200 GPUs. Second, it is optimized via dual-atomic RFT using GRPO \cite{shao2024deepseekmath} on a curated 1.4k dataset, performed on 40 NVIDIA H200 GPUs. The RFT dataset is curated using difficulty-aware rejection sampling. To \textit{efficiently} filter the initial candidate pool, we employ a coarse-grained assessment as a cost-effective proxy to assign a holistic quality score (on a 5-point scale) to each sample. 
We retain samples with scores in the empirically determined range of 2 to 4. For the RFT itself, where reward precision is paramount, the policy is guided exclusively by our fine-grained dual-atomic rewards.

\vspace{1ex}
\noindent\textbf{Data and Reward Generation Details.} 
Following common practice in recent image editing and reward-modeling works \cite{unifiedthinker,thinkrl,deepgen,endocot}, our data creation and reward computation process rely on a suite of high-performance off-the-shelf foundation models. We emphasize that these external models serve as implementation dependencies for data synthesis and reward provision, while the conceptual contribution of this work lies in the Thinker-centric training paradigm and dual-atomic reward design. For the initial data synthesis, we use Gemini-3-Pro to generate both the data tuples (source image description, user instruction, and rational reference description) based on our taxonomy and the expert-written target for SFT. Source images are synthesized from these source image descriptions using Gemini-2.5-Flash-Image. The checklists for our atomic reward mechanism are also generated by Gemini-3-Pro using the source image, user instruction, and rational reference description as input. To ensure focused and high-quality assessment, we constrain generation to a maximum of 6 questions for each checklist type (visual and cognitive). During the RFT stage, Gemini-3-Flash serves as the reward model within our dual-atomic reward framework to provide fine-grained feedback across both visual and cognitive dimensions.

\vspace{1ex}
\noindent\textbf{Evaluation Benchmarks.} 
We evaluate our model on two comprehensive benchmarks designed for reasoning-driven image editing. RISE-Bench \cite{risebench} presents complex reasoning challenges spanning temporal, causal, spatial, and logical axes. It evaluates the quality of the resulting edits based on three core metrics: instruction reasoning, appearance consistency, and visual plausibility. Complementing this, KRIS-Bench \cite{krisbench} provides a multifaceted assessment of factual, conceptual, and procedural knowledge, further organized into seven distinct cognitive dimensions.

\subsection{Main Results}

\vspace{1ex}
\noindent\textbf{Performance on RISE-Bench.} 
As shown in Table~\ref{tab:main results risebench}, when paired with Qwen-Image-Edit-2511, DDA-Thinker-32B achieves an overall accuracy of 40.0\%, the strongest performance among the reported community models on RISE-Bench. Our models also show substantial gains over the base Qwen-Image-Edit-2511 (19.4\%). We additionally report results on the Qwen-Image-Edit-2509 for a broader comparison. In this setting, our models again demonstrate strong performance, outperforming previous methods like Unified Thinker \cite{unifiedthinker} (28.9\%) and ThinkRL-Edit \cite{thinkrl} (29.7\%). This result is notable because our approach exclusively optimizes the Thinker, whereas strong prior methods either additionally update the Editor (e.g., Unified Thinker, ThinkRL-Edit) or rely on multi-round iterative refinement (e.g., EditThinker). The performance boost observed in DDA-Thinker-32B suggests that scaling the Thinker improves its ability to interpret complex instructions for visual transformation. This consistent scalability supports the efficacy of our dual-atomic RFT strategy. When evaluated against proprietary systems, DDA-Thinker-32B demonstrates competitive performance. It outperforms Gemini-2.5-Flash-Image (32.8\%) and achieves stronger temporal performance than Gemini-3-Pro-Image (45.9\% \textit{vs.}41.2\%). While a performance gap remains against the frontier Gemini-3-Pro-Image in broader causal and logical reasoning, our analysis suggests this is partly attributable to the limitations of the frozen Editor. The logical sub-task is particularly illustrative, as successfully rendering abstract concepts often pushes beyond the capabilities of current community editors. Despite this constraint, our model still achieves the second-best performance on this task among all community models. Importantly, these results should not be interpreted as evidence that decoupled optimization is superior to joint training.
Instead, they demonstrate that improving the planning module alone can already yield substantial gains and achieve high performance under a fixed editor setting.

\vspace{1ex}
\noindent\textbf{Performance on KRIS-Bench.} 
Table~\ref{tab:main results krisbench} shows that DDA-Thinker-32B reaches 79.94\% on KRIS-Bench with Qwen-Image-Edit-2511, outperforming all reported community models. We additionally report results on Qwen-Image-Edit-2509 for completeness. DDA-Thinker-32B (72.21\%) remains slightly above strong prior methods such as EditThinker (71.91\%) and ThinkRL-Edit (71.65\%). This comparison is noteworthy because our method uses a single-pass, frozen-editor setting, whereas EditThinker adopts a multi-round iterative refinement framework and ThinkRL-Edit updates the Editor during training. We observe a consistent scaling trend across both Editor settings, with DDA-Thinker-32B consistently outperforming DDA-Thinker-8B. This pattern suggests that our framework can effectively benefit from stronger Thinkers. This trend is consistent with the Thinker’s central role in planning. At the same time, the 2509 results also reveal a limitation: although both the 8B and 32B Thinkers substantially improve the overall score, some sub-dimensions remain slightly below the base editor. We attribute this to the limited capacity of the older frozen Editor, which may not fully realize more sophisticated plans. The decoupled setting makes this behavior easier to analyze by separating planning quality from editor execution. In a granular analysis against proprietary models, our DDA-Thinker-32B paired with Qwen-Image-Edit-2511 not only surpasses Gemini-2.5-Flash-Image (78.17\%) but also comes close to GPT-4o (80.09\%). It further outperforms GPT-4o in both Factual Knowledge (81.52\% vs. 79.80\%) and Procedural Knowledge (78.88\% vs. 78.32\%). While a performance gap exists relative to the frontier Gemini-3-Pro-Image (85.31\%), our model's strong performance supports the effectiveness of our Thinker-centric design in narrowing the gap between high-level instruction understanding and precise visual manipulation. 

\subsection{Ablation Study}
To validate the effectiveness of the key components in our DDA-Thinker, we conduct a series of ablation studies. Unless otherwise noted, all ablation experiments use our DDA-Thinker-32B and Qwen-Image-Edit-2511 on the challenging RISE-Bench dataset.

\begin{table}[t]
\centering
\caption{Ablation study on dual-atomic reward. The first row represents our thinker initialized via SFT.}
\label{tab:ablation dual-atomic}
\setlength{\tabcolsep}{1.5pt}
\fontsize{8pt}{12pt}\selectfont
\begin{tabular}{cc|ccc|c}
\toprule
Visual-Atomic & Cognitive-Atomic & Reason. & Consist. & Visual. & Overall Acc. \\ \midrule
     &      &   65.9    &     81.1  & 91.5 & 34.2  \\
$\checkmark$ &      &    66.8   &    83.5  & 91.8 & 36.6 \\
\rowcolor{gray!15}
$\checkmark$ & $\checkmark$ &   \textbf{68.4}    &    \textbf{84.4}   & \textbf{92.1}& \textbf{40.0} \\
\bottomrule
\end{tabular}
\end{table}

\begin{figure}[t]
\centering
\includegraphics[width=\linewidth]{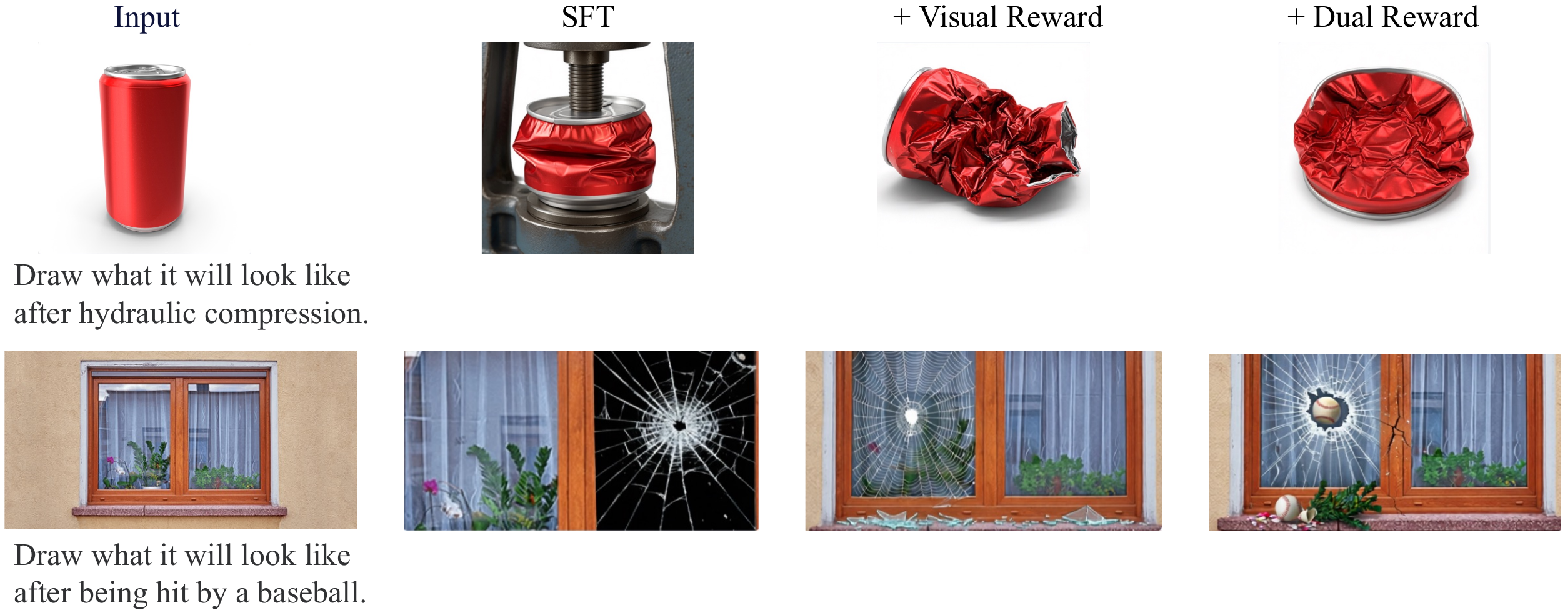}
\caption{\textbf{Qualitative Visualization of Our Reward Components.} 
The top row (`hydraulic press') shows the visual reward correcting the SFT model's hallucinated press to improve scene consistency.
The bottom row (`baseball') highlights the role of the cognitive reward in supporting physically grounded reasoning to render a faithful impact shatter.
}
\label{fig:ablation_study}
\end{figure}

\vspace{1ex}
\noindent\textbf{Effectiveness of Dual-Atomic Reward.} 
Table \ref{tab:ablation dual-atomic} reveals the impact of the proposed dual-atomic reward components. Compared to the SFT-initialized base (34.2\%), integrating visual-atomic reward improves the overall accuracy to 36.6\%. As evidenced by the table, this visual-atomic reward enhances appearance consistency (from 81.1\% to 83.5\%), as it leverages execution feedback from the final edited image. However, this purely visual-centric approach lacks direct feedback on the planning stage. By further incorporating cognitive-atomic reward, we achieve an additional performance gain to 40.0\%. Notably, this cognitive-atomic reward drives an improvement in instruction reasoning (from 66.8\% to 68.4\%), as it provides a more direct reward signal for plan quality. The synergy between these two dimensions confirms that, while visual feedback is essential for maintaining output quality, cognitive verification is necessary to reduce plan drift. 
Figure~\ref{fig:ablation_study} qualitatively visualizes this synergy: the visual reward corrects scene consistency (top), while the cognitive reward supports physically grounded reasoning to produce a faithful impact shatter (bottom). By bridging the gap between pixel-level generation and instruction-level planning, our dual-atomic framework helps ensure that generated plans remain both executable and logically coherent for complex tasks.

\vspace{1ex}
\noindent\textbf{Impact of Reward Granularity.} 
Table \ref{tab:ablation reward granularity} shows that our atomic-based reward mechanism achieves 40.0\%, outperforming the interval-based coarse scalar reward (36.9\%) \cite{unifiedthinker,dancegrpo,flowgrpo}. 
This performance gap highlights the limitations of using coarse scalar ratings for complex reasoning tasks, as they conflate multiple quality dimensions into a single ambiguous signal and lack the sensitivity to penalize subtle yet critical flaws. By contrast, our atomic-based approach decomposes holistic assessment into objective binary predicates that verify specific constraints. This shift from ambiguous scalar feedback to explicit and verifiable constraints provides the policy with clearer learning signals, helping the model remain aligned with physical logic and avoid visually plausible but logically inconsistent results.

\begin{table}[t]
\centering
\caption{Ablation study on reward granularity. Comparison between coarse scalar scores and our atomic-based reward mechanism. Both configurations use our dual-reward (visual + cognitive) reinforcement learning.}
\label{tab:ablation reward granularity}
\setlength{\tabcolsep}{5pt}
\fontsize{8pt}{12pt}\selectfont
\begin{tabular}{l|ccc|c}
\toprule
Method & Reason. & Consist. & Visual. & Overall Acc. \\
\midrule
coarse scalar reward  &  67.1 &  83.3 &  91.9 & 36.9  \\
\rowcolor{gray!15}
atomic-based reward  & \textbf{68.4} & \textbf{84.4} & \textbf{92.1} & \textbf{40.0}  \\
\bottomrule
\end{tabular}
\end{table}

\vspace{1ex}
\noindent\textbf{Impact of Reference-Grounded Checklists.} 
A design choice in our checklist synthesis is to ground checklist generation not only in the source image and user instruction but also in a rational reference description of the ideal post-edit scene. This grounding is intended to enhance checklist quality and reduce reward noise. To validate this design, we conduct an ablation study, presented in Table \ref{tab:ablation_rational_reference}. We compare our full DDA-Thinker model against a variant where checklists are generated without this rational reference description. Both models are trained using the dual-atomic (visual + cognitive) RL framework. Our reference-grounded strategy improves performance from 38.1\% to 40.0\%. Without this grounding, checklists are more susceptible to ambiguity in the user instruction, leading to less precise and potentially conflicting reward signals. By aligning the verification questions with the rational reference description of the ideal post-edit scene, our method supports more robust policy optimization, ultimately leading to superior planning and editing capabilities.

\begin{table}[t]
\centering
\caption{Ablation study on the impact of reference-grounded checklists. The only difference is whether the checklists are generated with the aid of a rational reference description.}
\label{tab:ablation_rational_reference}
\setlength{\tabcolsep}{5pt}
\fontsize{8pt}{12pt}\selectfont
\begin{tabular}{l|ccc|c}
\toprule
Checklist Generation & Reason. & Consist. & Visual. & Overall Acc. \\
\midrule
w/o rational reference & 67.5 & 83.2 & 92.0 & 38.1 \\
\rowcolor{gray!15}
w/ rational reference & \textbf{68.4} & \textbf{84.4} & \textbf{92.1} & \textbf{40.0} \\
\bottomrule
\end{tabular}
\end{table}

\vspace{1ex}
\noindent\textbf{Impact of Reward Integration Strategy.}
A natural alternative for integrating the visual and cognitive reward signals is to combine them into a single reward via a weighted sum and then apply standard GRPO on the merged reward. We compare this baseline (using equal weighting) against our design, which keeps the two rewards separate throughout optimization: each rollout group is associated with a single reward signal, and GRPO is applied to each group independently. Both configurations share identical data and dual-atomic rewards; they differ only in how the two reward signals are processed during optimization. As shown in Table~\ref{tab:ablation optimization strategy}, our design improves overall accuracy from 37.5\% to 40.0\%. We attribute this gap to two effects associated with the weighted-sum scheme. First, these two rewards encode semantically distinct dimensions (plan quality and visual quality) and often exhibit different within-group variances. When merged, the advantage is shaped by a single pooled standard deviation, allowing the objective with higher variance to overshadow the other and diminishing the relative gradient contribution of the more stable signal. Second, within a single rollout group, responses that excel in visual quality but fail in logic, as well as those that are logically sound but visually deficient, receive a moderate average reward. This suppresses the advantage value and weakens the policy-gradient signal when the two objectives are in conflict. In contrast, our design evaluates each objective independently within its own rollout group, ensuring that each reward signal provides a clear and well-scaled advantage to the subsequent policy update process.
    
\begin{table}[t]
\centering
\caption{Ablation study on reward integration strategy. Both configurations use the same dual-atomic rewards and differ only in how the visual and cognitive signals are integrated during optimization.}
\label{tab:ablation optimization strategy}
\setlength{\tabcolsep}{5pt}
\fontsize{8pt}{12pt}\selectfont
\begin{tabular}{l|ccc|c}
\toprule
Optimization Strategy & Reason. & Consist. & Visual. & Overall Acc. \\
\midrule
weighted sum   &  67.3 &  83.6 &  92.0 & 37.5  \\
\rowcolor{gray!15}
separate rewards  & \textbf{68.4} & \textbf{84.4} & \textbf{92.1} & \textbf{40.0}  \\
\bottomrule
\end{tabular}
\end{table}

\vspace{1ex}
\noindent\textbf{Impact of Difficulty-Aware Refinement.}
We introduce a difficulty-aware refinement pipeline to curate an effective training curriculum for reinforcement learning. To quantify the impact of this strategy, we conduct an ablation study comparing our model trained on the refined dataset against a baseline trained on the entire unrefined dataset. Both models use the same dual-atomic RFT framework and differ only in the training data. The results in Table~\ref{tab:ablation_difficulty_refinement} validate the effectiveness of this design. The model trained with difficulty-aware refinement data achieves an overall accuracy of 40.0\%, a substantial gain over the 36.7\% achieved on the unrefined data. This gain is observed across all dimensions, especially in appearance consistency. This improvement highlights the importance of data quality for RFT. The unrefined dataset that contains many trivial, overly difficult, or noisy samples may introduce conflicting gradients and hinder policy learning. By focusing training on a smaller yet more impactful set of moderately difficult samples, our refinement strategy provides a clearer and more concentrated learning signal. This result shows that our data curation pipeline is an important component of the DDA-Thinker rather than a negligible preprocessing detail.

\begin{table}[t]
\centering
\caption{Ablation on the impact of Difficulty-Aware Refinement. We compare RFT on the curated refined dataset against training on the full unrefined dataset.}
\label{tab:ablation_difficulty_refinement}
\setlength{\tabcolsep}{3pt}
\fontsize{8pt}{12pt}\selectfont
\begin{tabular}{l|ccc|c}
\toprule
RFT Training Data & Reason. & Consist. & Visual. & Overall Acc. \\
\midrule
w/o difficulty-aware refinement &  67.8 &  81.7 &  91.6 & 36.7  \\
\rowcolor{gray!15}
w/ difficulty-aware refinement  & \textbf{68.4} & \textbf{84.4} & \textbf{92.1} & \textbf{40.0}  \\
\bottomrule
\end{tabular}
\end{table}

\begin{figure*}[ht]
\centering
\includegraphics[width=\textwidth]{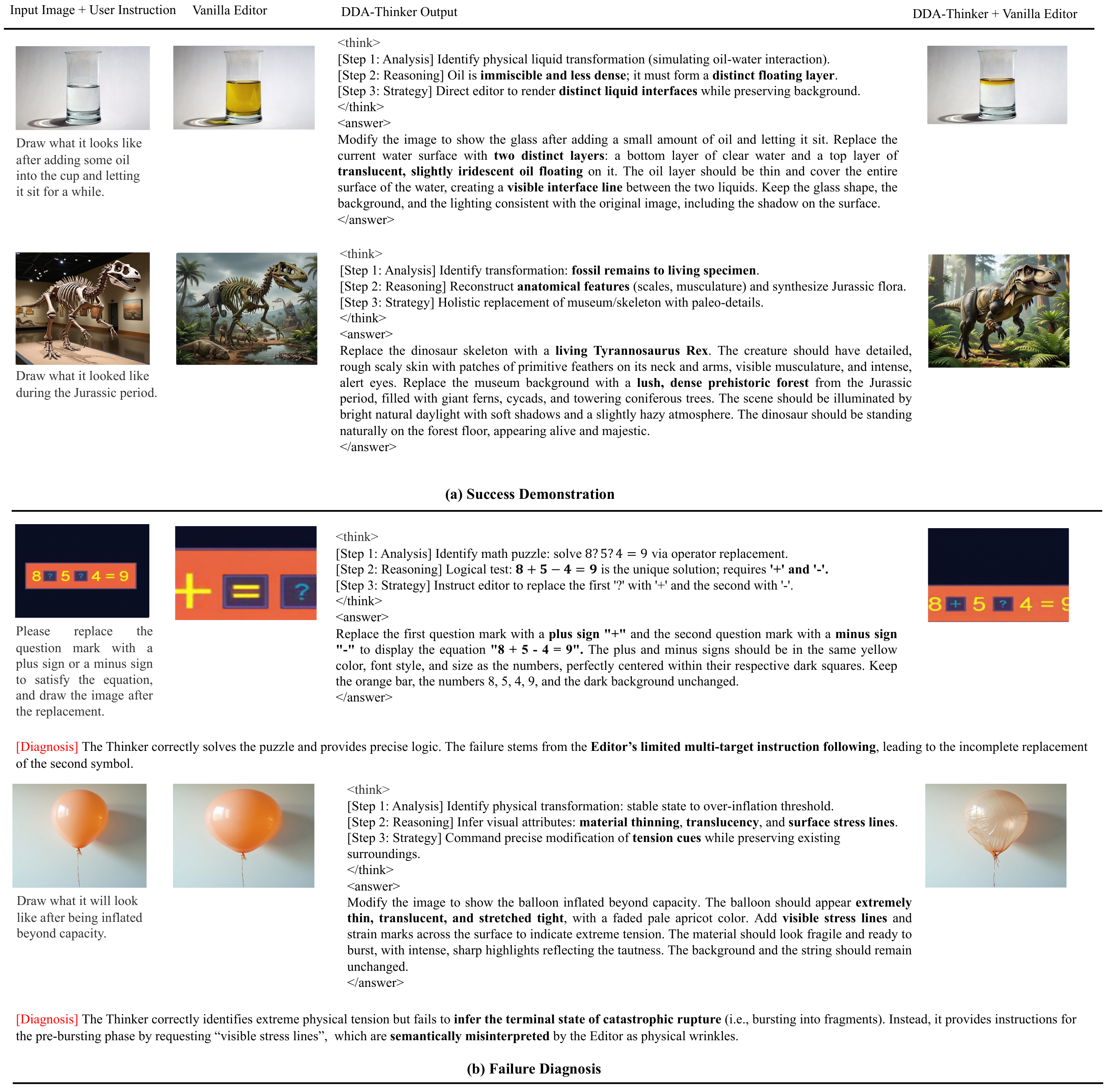} 
\caption{\textbf{Visual Comparison of Results Without and With DDA-Thinker.}
(a) Success demonstration of reasoning-driven image editing. (b) Failure diagnosis of system boundaries.
\textit{Note}: \textless think\textgreater \ blocks are condensed for readability; \textless answer\textgreater \ blocks are provided in full.}
\label{fig:qualitative_results}
\end{figure*}

\vspace{1ex}
\noindent\textbf{Zero-shot Thinker Transferability.} 
Table \ref{tab:ablation cross-architecture transferability} highlights the zero-shot transferability of our DDA-Thinker, evaluating its effectiveness as a planning module for unseen editors. When paired with FLUX.1-Kontext-Dev\cite{blackforest2024flux}, our DDA-Thinker improves the overall accuracy from 5.8\% to 18.9\%. Similarly, it improves LongCat-Image-Edit\cite{longcat} from 12.8\% to 27.8\%. These substantial gains are achieved without any parameter updates or target-specific retraining. The significant improvements in instruction reasoning and appearance consistency metrics across different architectures suggest that our model has learned transferable planning patterns. This successful transfer supports the portability of our design and suggests that the learned Thinker can function as a reusable planning module for other editors. It also suggests that the planning patterns learned during training remain useful when transferred to other editors, offering a scalable pathway for enhancing existing generative models with stronger planning capabilities.

\begin{table}[t]
\centering
\caption{Zero-shot transferability to other editors without further training our DDA-Thinker.}
\label{tab:ablation cross-architecture transferability}
\setlength{\tabcolsep}{5pt}
\fontsize{8pt}{12pt}\selectfont
\begin{tabular}{l|ccc|c}
\toprule
Method & Reason. & Consist. & Visual. & Overall Acc. \\
\midrule
FLUX.1-Kontext-Dev &  26.0 & 71.6 & 85.2 & 5.8  \\
\rowcolor{gray!15}
{\scriptsize\hspace{2pt} + DDA-Thinker-32B} & \textbf{52.8} & \textbf{77.1} & \textbf{86.9} & \textbf{18.9} \\
\midrule
LongCat-Image-Edit & 38.5  & 76.2 & 88.7 &  12.8 \\
\rowcolor{gray!15}
{\scriptsize\hspace{2pt} + DDA-Thinker-32B} & \textbf{63.7} & \textbf{82.4} &  \textbf{90.2} & \textbf{27.8} \\
\bottomrule
\end{tabular}
\end{table}

\vspace{1ex}
\noindent\textbf{Comparison with Proprietary Foundation Planning Models.} 
Table \ref{tab:ablation foundation models} compares DDA-Thinker-32B with proprietary foundation planning models when integrated with Qwen-Image-Edit-2511 \cite{qwenimage-edit}. 
DDA-Thinker-32B achieves an overall accuracy of 40.0\%, outperforming Gemini-2.5-Pro (38.3\%). While Gemini-3-Pro maintains a lead in the instruction reasoning metric (72.5\%), DDA-Thinker-32B demonstrates superior appearance consistency (84.4\%) and visual plausibility (92.1\%). This performance profile is consistent with our dual-atomic reinforcement learning strategy: the visual-atomic reward helps ground the executable plan in the Editor's synthesis constraints, while the cognitive-atomic reward improves logical quality at the plan level. Together, these mechanisms help align the Thinker’s planning output with the Editor’s execution capacity, thereby effectively enabling stronger performance on consistency-sensitive tasks.

\begin{table}[t]
\centering
\caption{Comparison with proprietary foundation planning models.}
\label{tab:ablation foundation models}
\setlength{\tabcolsep}{4pt}
\fontsize{8pt}{12pt}\selectfont
\begin{tabular}{l|ccc|c}
\toprule
Method & Reason. & Consist. & Visual. & Overall Acc. \\
\midrule
Qwen-Image-Edit-2511 &  49.9 & 71.0 & 91.5 & 19.4  \\
\midrule
{\scriptsize\hspace{2pt} + Gemini-3-Pro} & \textbf{72.5} & 83.4 & 91.7 & \textbf{41.7}\\
{\scriptsize\hspace{2pt} + Gemini-2.5-Pro} & 71.0 & 78.9 & 91.5 & 38.3 \\
\rowcolor{gray!15}
{\scriptsize\hspace{2pt} + DDA-Thinker-32B} & 68.4 & \textbf{84.4} & \textbf{92.1} & 40.0 \\
\bottomrule
\end{tabular}
\end{table}

\subsection{Qualitative Analysis}
Figure~\ref{fig:qualitative_results} demonstrates the capabilities of the vanilla editor (Qwen-Image-Edit-2511) when augmented with DDA-Thinker. It presents both success cases that highlight the improved reasoning-driven image editing and failure cases that reveal the current boundaries of the system.

\vspace{1ex}
\noindent\textbf{Success Demonstration.} 
Figure~\ref{fig:qualitative_results}(a) shows that DDA-Thinker elevates the vanilla editor from performing superficial edits to executing more complex reasoning-grounded transformations. 
This is evident in the dinosaur restoration task, where the vanilla editor fails by simply placing the fossil in a generic forest, while DDA-Thinker provides precise biological grounding to guide a holistic transition from skeleton to a fleshed-out organism within a plausible Jurassic habitat. 
The same pattern holds for physical reasoning. In the ``oil-water" case, the vanilla editor incorrectly renders a physically implausible homogenous mixture. In contrast, DDA-Thinker's cognitive plan captures the principles of immiscibility and density, directing the Editor to create a distinct floating oil layer that better adheres to physical common sense. In both scenarios, DDA-Thinker enables edits that the Editor alone does not reliably realize.

\vspace{1ex}
\noindent\textbf{Failure Diagnosis.} 
Figure~\ref{fig:qualitative_results}(b) illuminates the system's current boundaries, where our approach enables more precise failure attribution. In the ``math puzzle" example, the failure is isolated to the Editor: while DDA-Thinker correctly solves the logic ($8+5-4=9$), the Editor fails to execute the second symbol replacement due to ``multi-target neglect." The ``balloon" case, by contrast, reveals a limitation in the Thinker's cognitive plan, which focuses on pre-bursting tension instead of the terminal state of rupture. These cases illustrate that, while DDA-Thinker improves planning quality, bottlenecks remain in both the Editor’s execution capabilities and the completeness of the generated executable plan.
\section{Conclusion}
We present DDA-Thinker, a Thinker-centric framework for reasoning-driven image editing that decouples planning from generation to explicitly optimize the planning module under a frozen Editor, thereby enabling a more controlled analysis of how improving the Thinker affects editing performance. In this setting, executable planning is the form through which the Thinker’s reasoning becomes directly usable by the Editor. To support this optimization, we introduce a dual-atomic reinforcement learning framework with a cognitive-atomic reward and a visual-atomic reward, both implemented through checklist-based verification. To further improve checklist quality, their synthesis is further grounded in a rational reference description of the ideal post-edit scene. Together, these designs enable our approach to produce more accurate, executable, and logically consistent editing plans, ultimately leading to higher-quality edited images. Extensive experiments further show that explicitly optimizing the Thinker can substantially improve reasoning-driven image editing without modifying the underlying generator, offering a promising direction while remaining complementary to existing joint optimization paradigms rather than replacing them.

\section{Limitations and Future Work}
Our study deliberately focuses on a decoupled Thinker-centric paradigm to investigate the impact of an improved Thinker on a fixed Editor. While our results demonstrate the potential of this approach, we acknowledge two primary avenues for future improvement.

First, our paradigm of optimizing only the Thinker, while methodologically well motivated, means that performance is ultimately bounded by the capabilities of the frozen Editor, which can become a bottleneck. This is most apparent in tasks that demand the generation of highly symbolic content, such as mathematical equations or logical notation. Here, a strong executable plan may still fail if the Editor lacks the ability for precise and structured rendering. A natural future direction is to explore hybrid optimization schemes that selectively fine-tune the Editor for such tasks. However, the primary challenge is to implement such a hybrid approach without compromising the core benefits of our framework: the clearer credit assignment enabled by decoupling and the precise feedback supported by dual-atomic rewards.

Second, the reliability of our reward generation pipeline, from synthesizing dual-atomic checklists to final reward evaluation, is dependent on the capabilities of external proprietary foundation models. These models can introduce subtle noise or biases into the guidance signals, potentially creating a performance ceiling. This dependency motivates future research on more robust reward mechanisms that reduce reliance on external proprietary model calls for data generation and reward evaluation. Accordingly, the current results should be interpreted in the context of a Thinker-centric optimization pipeline that uses external foundation models as implementation tools, rather than a fully self-contained training pipeline.

\bibliographystyle{IEEEtran}
\bibliography{main}


\end{document}